\documentclass{article}

\usepackage{PRIMEarxiv}

\usepackage[utf8]{inputenc}
\usepackage[T1]{fontenc}

\usepackage{amsmath}
\usepackage{amssymb}
\usepackage{amsfonts}
\usepackage{physics}
\usepackage{orcidlink}

\usepackage{graphicx}
\usepackage{float}
\usepackage{caption}
\usepackage{subcaption}

\usepackage{booktabs}
\usepackage{tabularx}
\usepackage{multirow}
\usepackage{adjustbox}

\usepackage{algorithm}
\usepackage{algpseudocode}

\usepackage{microtype}
\usepackage{textcomp}
\usepackage{nicefrac}
\usepackage{siunitx}

\usepackage{url}

\usepackage{geometry}
\usepackage{fancyhdr}

\usepackage{cite}
\usepackage{lipsum}
\usepackage{pdfpages}
\usepackage{array}

\newcolumntype{M}[1]{>{\centering\arraybackslash}m{#1}}

\usepackage{hyperref}

\title{Physiologically Grounded Driver Behavior Classification: SHAP-Driven Elite Feature Selection and Hybrid Gradient Boosting for Multimodal Physiological Signals}

\author{
Sahar Askari \\
Department of Computer Science and Engineering \\
Shiraz University \\
Shiraz, Iran \\
\texttt{askarisahar92@gmail.com} \\
\And
Mohammad Mahdi Mirza Ali Mohammadi \text{\orcidlink{0009-0000-4662-6031}} \\
Department of Electrical Engineering \\
Iran University of Science and Technology \\
Tehran, Iran \\
\texttt{hamed.m.a.mohammadi@gmail.com} \\
\And
Fatemeh Ensafdoust \\
Department of Computer Engineering \\
Islamic Azad University, Rasht Branch \\
Rasht, Iran \\
\texttt{ensafdoust@gmail.com} \\
\And
Amin Golnari \text{\orcidlink{0000-0002-3841-0588}} \thanks{Corresponding author: Amin Golnari \texttt{<amingolnarii@gmail.com>}} \\
Faculty of Electrical Engineering \\
Shahrood University of Technology \\
Shahrood, Iran \\
\texttt{amingolnarii@gmail.com} \\
\And 
Saeid Sanei \text{\orcidlink{0000-0002-1446-5744}} \\
College of Engineering and Computer Science \& School of Neuroscience \\
VinUniversity \& King’s College London \\
Hanoi, Vietnam \& London, United Kingdom \\
\texttt{saeid.s@vinuni.edu.vn}
}

\begin{document}

\maketitle

\begin{abstract}
An interpretable and scalable framework for decoding driving behaviors from multimodal physiological signals is proposed in this study. We utilize multimodal physiological driving behavior large-scale dataset comprising synchronized electroencephalogram (EEG), electromyography (EMG), and galvanic skin response (GSR) signals. Our approach involves rigorous preprocessing followed by a domain-specific feature extraction pipeline targeting time-domain, frequency-domain, and derived physiological indices. To address high dimensionality, we employ SHAP-based elite feature selection, retaining the top 250 features to reduce computational overhead while preserving predictive power. Hyperparameter optimization for extreme gradient boosting (XGBoost) and light gradient boosting machine (LightGBM) models is conducted using Bayesian optimization via Optuna. Finally, a weighted soft-voting ensemble is constructed to leverage the complementary strengths of both gradient boosting frameworks. The results demonstrate that the proposed ensemble achieves a test accuracy of 80.91\% and a macro-F1 score of 0.79, significantly outperforming single-modality baselines and traditional machine learning models. Ablation studies confirm a 8\% performance gain over the best single modality (EEG), validating the necessity of multimodal fusion. SHAP analysis further validates the physiological plausibility of the model, revealing that the EEG contributes the majority of predictive weight, GSR and EMG features provide critical discriminatory signals for high-arousal and motor-intensive maneuvers.
\end{abstract}


\section{Introduction}

Road traffic injuries remain a critical global public health concern, accounting for approximately 1.19 million fatalities annually worldwide \cite{world2024global}. Approximately 94\% of traffic crashes are implicated by human factors, which include distraction, inattention, fatigue, and inappropriate driving actions \cite{singh2015critical,dingus2016driver}, underscoring the urgent need for intelligent driver-monitoring systems capable of identifying unsafe behaviors before they escalate into critical events. As modern vehicles increasingly incorporate automation and advanced driver-assistance systems, understanding and modeling the cognitive and physiological mechanisms underlying driver behavior has become fundamental to improving road safety, human–vehicle interaction, and autonomous decision-making.

Conventional driver-monitoring approaches predominantly rely on vehicle kinematics, such as steering-wheel variability, lane deviation, acceleration patterns, and speed fluctuations \cite{liu2020differences,xing2020driver,bi2015detecting}. While effective in detecting overt behavioral deviations, these measures primarily reflect the outcomes of cognitive failures rather than the underlying cognitive and physiological processes that precede them. Vision-based techniques, including eye tracking and head-pose estimation, have further advanced the assessment of visual attention, gaze allocation, and fatigue-related behaviors \cite{7478592,ahlstrom2021eye,michelaraki2023real}. Nevertheless, such external measures remain limited in their ability to capture internal perceptual, attentional, and decision-making dynamics that directly drive changes in driving behavior \cite{di2018eeg}.

To address these limitations, increasing attention has been directed toward physiological and neurophysiological signals, particularly electroencephalography (EEG), which provides millisecond-level temporal resolution of neural activity \cite{sanei2021eeg}. EEG oscillations across theta, alpha, and beta frequency bands have been strongly associated with cognitive control, vigilance, attentional allocation, motor preparation, and mental fatigue \cite{klimesch2012alpha,cavanagh2014frontal,getzmann2018age,alyan2023operator}. Anticipatory neural components, such as the contingent negative variation (CNV) and movement-related cortical potentials, precede braking, steering, and decision-making actions by hundreds of milliseconds, enabling early inference of driver intentions \cite{nguyen2019detection}. Consequently, a growing body of research has demonstrated the feasibility of decoding lane changes, emergency braking, workload, distraction, and fatigue from EEG signals \cite{kim2014detection,zhang2024}.

Despite its advantages, EEG alone is constrained by limited spatial resolution and susceptibility to noise and motion artifacts, particularly in real-world driving environments \cite{lorenz2018dissociating,bansal2019cognitive}. This has motivated the development of multimodal driver-monitoring systems that integrate EEG with complementary physiological signals such as electromyography (EMG) and galvanic skin response (GSR). Each modality captures distinct yet complementary aspects of driver state: EMG reflects muscle activation patterns, while GSR indexes arousal and stress responses. Although other modalities like electrocardiography (ECG) are also utilized in the literature, the combination of EEG, EMG, and GSR offers a robust profile of cognitive, motor, and autonomic states \cite{sahayadhas2012detecting}. Empirical evidence consistently shows that multimodal fusion substantially outperforms unimodal approaches in fatigue detection, emotion recognition, cognitive-state decoding, and intention prediction \cite{lian2024multimodal,10575932,dontoh2025visual}.

Recent large-scale multimodal datasets, such as manD 1.0 \cite{DargahiNobari2024}, have enabled more comprehensive modeling of driver behavior across varying task demands and automation levels. However, many existing datasets remain limited by modality richness, temporal alignment, or behavioral diversity \cite{rito2020neurobehavioural}. To address these gaps, Tao et al. (2024) introduced the multimodal physiological driving behavior (MPDB) dataset \cite{tao2024multimodal}. Baseline evaluations using linear discriminant analysis (LDA), convolutional and recurrent neural networks, EEGNet, and multimodal fusion architectures demonstrated that integrating multiple physiological modalities significantly improves behavior classification accuracy, highlighting the strong coupling between physiological responses and driving actions.

While deep learning approaches have shown promising results on multimodal physiological data, they often require large amounts of labeled data, are computationally demanding, and offer limited interpretability. These factors hinder deployment in safety-critical and resource-constrained environments. In contrast, classical machine learning methods, particularly tree-based ensemble models such as extreme gradient boosting (XGBoost) \cite{chen2016xgboost} and light gradient-boosting machine (LightGBM) \cite{ke2017lightgbm}, provide strong performance on structured data, robustness to noise, efficient training, and inherent interpretability through feature importance and explainability techniques. Their effectiveness, however, depends critically on the design of informative and physiologically meaningful feature representations.

In this work, we propose an interpretable and scalable framework for decoding driving behaviors from multimodal physiological signals in the MPDB dataset. The MPDB raw dataset provides synchronized 64-channel EEG, 4-channel EMG, and GSR signals collected from 30 participants, offering a rich source of neural, muscular, and autonomic data. Our approach begins with a rigorous preprocessing pipeline that includes band-pass filtering, independent component analysis (ICA) for artifact removal, and precise multimodal synchronization. For each modality, we extract a comprehensive set of features, including time-domain metrics (e.g., line length, variance of derivatives), frequency-domain power spectral density (PSD) across physiologically relevant bands, and derived indices such as the EEG Alpha/Theta ratio and EMG asymmetry.

Feature extraction is executed in a streaming manner, enabling memory-efficient processing and seamless scalability to large datasets. To address the high dimensionality of the feature space, we employ a SHAP-based (SHapley Additive exPlanations) elite feature selection strategy. An auxiliary XGBoost model is used to compute global feature importance, allowing us to retain the top 250 most influential features while preserving complex nonlinear interactions. Final classification is performed using a weighted soft-voting ensemble of XGBoost and LightGBM models, optimized via Bayesian optimization. This hybrid approach leverages the complementary strengths of both gradient boosting frameworks to achieve robust performance on the four-class driving behavior problem, including Brake, Change (Lane Change), Throttle (Acceleration), and Turn.

By combining physiologically informed feature engineering, advanced artifact removal, and explainable tree-based ensemble learning, the proposed framework offers a practical and transparent alternative to end-to-end deep learning approaches for multimodal driver-behavior recognition. It is particularly well suited for scenarios where data efficiency, interpretability, and robustness are essential, and where early detection of unsafe driving behaviors can contribute meaningfully to real-time driver assistance and accident prevention.

The main contributions of this work are summarized as follows:

\begin{itemize}
    \item A robust preprocessing and feature extraction pipeline for the MPDB dataset, incorporating ICA-based artifact removal, multimodal synchronization, and the extraction of time-frequency and domain-specific physiological features.
    \item A subject-aware normalization and streaming processing strategy that reduces inter-subject bias and enables efficient handling of large-scale multimodal datasets.
    \item A SHAP-based elite feature selection approach that reduces dimensionality by 50\% while retaining critical predictive information, enhancing both model efficiency and interpretability.
    \item A hybrid ensemble learning framework combining XGBoost and LightGBM, optimized via Bayesian methods, which achieves state-of-the-art accuracy (80.91\%) and macro-F1 (0.79) on the MPDB dataset.
    \item A comprehensive interpretability analysis using SHAP values, validating the physiological plausibility of the model by highlighting the distinct contributions of EEG (cognitive load), EMG (motor execution), and GSR (arousal) to behavior classification.
\end{itemize}

\section{Background}

Physiological signals have emerged as indispensable indicators of driver cognitive and affective states, providing direct access to internal neurophysiological processes that precede observable behavioral manifestations. The sympathetic nervous system response to stressful driving conditions produces observable changes in cardiovascular activity, electrodermal activity, muscular tension, and neural activity, providing objective markers for continuous monitoring \cite{papakostas2021understanding}. Blood volume pulse (BVP) and heart rate variability (HRV) features, including inter-beat interval (IBI) statistics such as pNN50 and spectral power distributions, demonstrate strong associations with cognitive inattention, distraction, and drowsiness. These physiological markers often outperform visual modalities in detecting covert mental states that lack overt physical correlates. The assessment of driver states is further enriched by complementary physiological channels where respiration patterns show significant discriminative power for driver emotion state classification with an accuracy of 94.88\% while electrodermal activity (EDA) and EEG track emotional arousal dynamics and mental load through a hybrid feature selection algorithm \cite{huang2024driver}. Critically, multimodal fusion strategies consistently surpass unimodal approaches. Recent reviews confirm that combining modulities provides complementary information, with multimodal systems significantly outperforming single-modality baselines in stress and cognitive load detection. For instance, integrated physiological-visual systems have achieved up to 96\% F1-score in distraction recognition and 96\% accuracy in drowsiness detection \cite{amin2023real}.

Validation under ecologically valid conditions has further solidified the role of physiological sensing in driver monitoring. Large-scale naturalistic driving studies encompassing over 800 hours of real-world data have revealed significant correlations between vehicle kinematics (including speed fluctuations, acceleration patterns, and steering behavior) and drivers' physiological responses, particularly HRV and EDA. These findings demonstrate that physiological signatures systematically vary across traffic conditions and in-vehicle interactions \cite{milardo2022understanding}. Such findings have catalyzed adaptive monitoring frameworks like DrivNet, which correlates driver reactions, vehicle parameters, and environmental context to classify safe versus unsafe driving modes with 88.3\% accuracy while explicitly modeling psychological states including mental workload, stress, and fatigue \cite{theivadas2026drivnet}. Ecologically valid emotion induction protocols, such as Wizard-of-Oz experiments capturing natural affective states including frustration and anxiety, have further confirmed that emotional expressions manifest consistently across audio, video, and physiological modalities, enabling reliable annotation using both dimensional and categorical approaches \cite{lotz2018recognizing}. Optimal placement of wearable sensors, such as on the spine, has been shown to effectively discriminate motion patterns during cyclic activities \cite{prochazka2021discrimination}. Nevertheless, real-world deployment faces persistent challenges including sensor failures, motion artifacts, and transient signal loss. Emerging techniques in missing-modality learning now enable robust inference despite partial data availability, a critical capability for safety-critical applications requiring uninterrupted monitoring despite sensor degradation or occlusion \cite{wu2026deep}. As discussed in \cite{sanei2020body}, body sensor networks provide a framework for integrating multiple physiological measurements including neural, muscular, and autonomic signals through synchronized acquisition and cooperative processing architectures.

Methodological advances in computational modeling have substantially enhanced the capacity to extract meaningful patterns from complex multimodal physiological data. The field has evolved from classical machine learning with handcrafted features to sophisticated deep learning architectures. Early pioneering work demonstrated that skin conductivity and heart rate metrics correlate closely with observable stressors, achieving over 97\% accuracy in distinguishing stress levels using classical analysis \cite{healey2005detecting}. Contemporary approaches increasingly leverage deep learning, including convolutional, recurrent, and transformer-based networks. For example, hypergraph convolutional networks (HGCNs) model higher-order interactions among physiological channels beyond pairwise relationships, enabling unified representation of EEG, EMG, and GSR signals for driver intention recognition \cite{zhang2024mmphgcn}. Similarly, dual-branch architectures integrating spectral power features with functional connectivity metrics, such as phase-locking value (PLV), achieve over 90\% accuracy in mental workload classification while maintaining computational efficiency suitable for in-vehicle deployment \cite{liu2025eeg}. 

Notably, transformer-based architectures have recently demonstrated exceptional performance in this domain. The Phy-FusionNet \cite{wu2025phy}, a memory-augmented transformer with periodicity and contextual attention mechanisms, achieved accuracies up to 88.6\% on the CL-Drive dataset \cite{angkan2024multimodal} for cognitive load classification and 99.1\% on WESAD dataset \cite{schmidt2018introducing} for stress classification, highlighting the potential of self-attention mechanisms to capture complex temporal dependencies in physiological data. Transfer learning frameworks further address substantial inter-subject variability in physiological signals by aligning subject-specific distributions with generalized models using minimal calibration data, enabling rapid adaptation without extensive retraining \cite{lin2025hatnet}. For instance, deep transfer learning using pre-trained CNNs on ECG scalograms has achieved validation accuracies exceeding 98\% in stress detection tasks \cite{amin2022ecg}. Hybrid feature engineering strategies combining handcrafted time- and frequency-domain features with deep neural representations will be refined using explainability techniques such as SHAP. This refinement may enhance a proposed hybrid CNN-LSTM model that integrates multimodal physiological signals (heart rate, respiration, skin conductance) to monitor driver fatigue and stress, achieving 86.38\% accuracy \cite{zhou2024monitoring}. Modeling EEG channels as a cooperative network has proven effective for decoding fine motor movements, even with limited training data, by capturing neural interactions during prolonged tasks \cite{falcon2025cooperative}.

Despite these advances, the widespread deployment of deep learning models in safety-critical transportation systems remains constrained by their limited interpretability and transparency. The black-box nature of deep neural networks poses significant barriers to regulatory approval, trust, and real-world adoption, particularly in applications where understanding the reasoning behind predictions is essential for safety validation and human oversight \cite{al2026advances}. This has catalyzed growing interest in interpretable machine learning frameworks that balance predictive performance with transparency. Tree-based ensemble methods, including XGBoost and LightGBM, have emerged as particularly effective solutions for structured physiological data, offering strong predictive accuracy while providing intrinsic interpretability through feature importance analysis and model-agnostic explanation techniques such as SHAP \cite{liu2025high}. SHAP analysis enables both global interpretation of feature importance and local explanation of individual predictions, facilitating causal insight into how physiological markers, including EEG spectral power and heart rate variability features, contribute to driver state classification \cite{hussain2025explainable}. Empirical studies have demonstrated that SHAP-guided analysis can reveal meaningful patterns, such as identifying frontal theta activity and parietal alpha suppression as dominant indicators of elevated mental workload, or highlighting specific contributing factors (acceleration, speed, roll speed, pitch speed, and engine speed) associated with unsafe or inefficient driving behaviors \cite{liu2025high}. Importantly, interpretability also enhances user trust and system acceptance, as drivers and operators are more likely to rely on systems whose decisions align with established physiological principles and domain knowledge.

Tree-based ensemble learning has demonstrated particular effectiveness in safety-critical driving applications due to its robustness, computational efficiency, and explainability. Optimization-enhanced XGBoost frameworks achieve state-of-the-art performance in lane-change decision prediction, with SHAP analysis identifying longitudinal acceleration, relative vehicle speed, and lateral positioning as dominant predictors of driver intention \cite{ccetinkaya2023driver,sekadakis2025analyzing}. Similar explainable ensemble approaches have quantified how vehicle dynamics, obstacle proximity, and environmental conditions influence driver takeover behavior in automated driving scenarios, providing actionable insights into human–vehicle interaction dynamics \cite{sekadakis2025analyzing}. Moreover, stacking ensemble architectures combining XGBoost, LightGBM, and CatBoost with neural meta-learners achieve significant performance improvements in multimodal physiological state classification while preserving interpretability through SHAP-based analysis \cite{huang2024multilayer}. Classical methods such as support vector machines (SVM) and quadratic discriminant analysis (QDA) also remain relevant, with QDA achieving up to 96\% accuracy in binary stress classification using ECG and skin conductance features, demonstrating that carefully engineered features combined with efficient classifiers can rival more complex deep learning models in specific contexts \cite{marcantoni2022identification}. These findings establish tree-based ensemble methods not only as competitive alternatives to deep learning but as uniquely suited approaches for physiological driver monitoring, where transparent reasoning, computational efficiency, and robustness to noisy multimodal sensor data are essential for real-time deployment.

Collectively, these advances highlight the importance of integrating physiologically grounded feature representations with interpretable and computationally efficient machine learning models. Such approaches provide a practical pathway toward deployable driver monitoring systems capable of delivering accurate, transparent, and reliable behavior recognition under real-world operational constraints.

\section{Methodology}

\subsection{Dataset and Preprocessing}
The study utilizes the MPDB dataset \cite{tao2024multimodal}, which comprises synchronized physiological signals acquired during simulated driving tasks. The raw data includes 64-channel EEG, 4-channel EMG, and single-channel GSR. According to the dataset specifications, all physiological signals were originally acquired at a sampling frequency of 1000 Hz.

Unlike continuous streaming data, the MPDB dataset is structured in an event-related manner. Specific driving behaviors are annotated using time markers synchronized with the driving simulator. In this study, for each event, a temporal window of 2.0 seconds is extracted, spanning from $t = -0.5$ s (pre-event baseline) to $t = +1.5$ s (post-event response) relative to the event onset. To reduce computational complexity while preserving the relevant physiological frequency content, we downsampled all signals from 1000 Hz to 500 Hz. Consequently, each processed sample consists of 1001 discrete time points across 69 channels. 

The experiment employs a significantly larger, stratified dataset containing a total of 17,882 samples. The dataset is stratified partitioned into a 80\% training set ($n=14,305$) and a 20\% test set ($n=3,577$), ensuring proportional class distribution to evaluate model robustness on unseen data. The four targeted driving behavior classes are Brake, Change, Throttle, and Turn. Table \ref{tab:class_dist_new} details the class distribution.

\begin{table}[h]
\centering
\caption{Dataset Class Distribution.}
\label{tab:class_dist_new}
\footnotesize
\begin{tabular}{lcccc}
\toprule
Class & Training Samples & Test Samples & Total & Percentage \\
\midrule
Brake & 5,130 & 1,283 & 6,413 & 35.9\% \\
Change & 2,506 & 626 & 3,132 & 17.5\% \\
Throttle & 1,957 & 489 & 2,446 & 13.7\% \\
Turn & 4,712 & 1,179 & 5,891 & 32.9\% \\
\midrule
Total & 14,305 & 3,577 & 17,882 & 100\% \\
\bottomrule
\end{tabular}
\end{table}

A rigorous preprocessing pipeline was applied to ensure signal quality and multimodal synchronization:
\begin{itemize}
    \item \textbf{EEG Processing:} Raw 64-channel EEG data were band-pass filtered (0.5-40 Hz, IIR). Samples containing bad channels, identified via robust Z-score (MAD-based) and flatline detection, were excluded from further analysis. Artifact removal was performed using independent component analysis (ICA, Infomax algorithm) fitted on high-pass filtered data (1.0 Hz). Five non-cerebral channels (ECG, horizontal and vertical electrooculography) were excluded from the 64-channel EEG cap, resulting in 59 EEG channels for analysis, after which a standard 10-20 montage was applied.
    \item \textbf{EMG Processing:} Signals were band-pass filtered (20-240 Hz, IIR) and subjected to notch filtering at [50, 100, 150, 200] Hz (FIR) to eliminate power-line interference.
    \item \textbf{GSR Processing:} Data were band-pass filtered (0.1-35 Hz, IIR) to isolate phasic and tonic components.
    \item \textbf{Synchronization \& Fusion:} Multimodal alignment was achieved via hybrid cross-modality matching using a 3000ms temporal window and strict Event-ID verification. Jitter was managed through nearest-neighbor timestamp matching. Baseline correction was applied using the pre-event period (-0.5s to 0.0s). Late fusion was implemented via channel-axis concatenation, with automated truncation to the minimum common epoch count to maintain data integrity.
\end{itemize}

\begin{figure}[h]
\begin{center}
\includegraphics[width = 16 cm, clip = true, trim = 0 0 0 0]{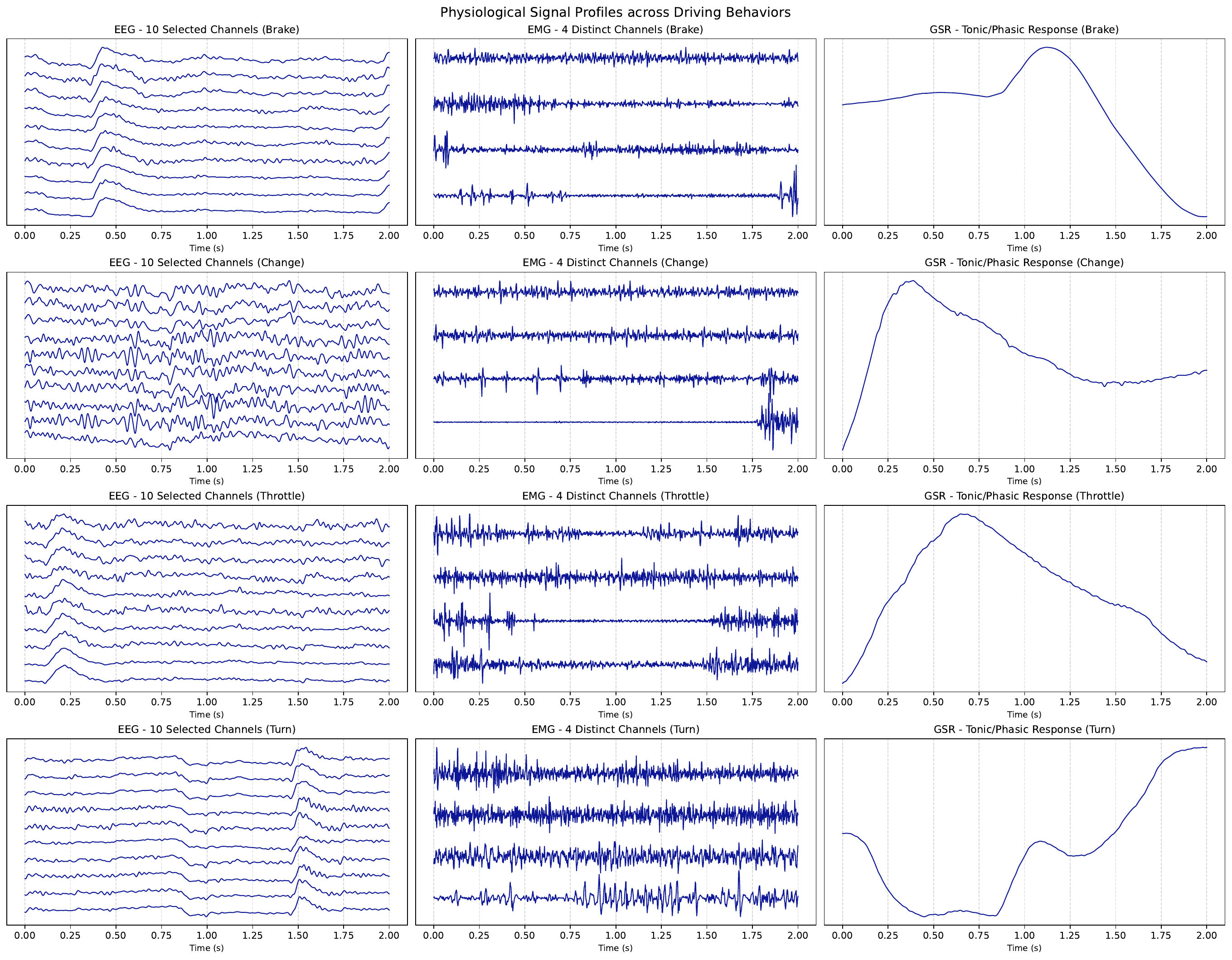}
\caption{Representative 2-second windows of synchronized multimodal physiological signals across the four driving behavior classes. The visualization showcases the first 10 EEG channels (stacked), 4 EMG channels (separated by offset), and the single-channel GSR response, illustrating the distinct temporal dynamics and signal integrity achieved after the ICA-based artifact removal and preprocessing pipeline.}
\label{fig:multimodal_samples}
\end{center}
\end{figure}

All normalization parameters and feature selection statistics were derived exclusively from the training partition to prevent information leakage. Figure \ref{fig:multimodal_samples} illustrates representative physiological signal profiles for EEG, EMG, and GSR modalities across the four targeted driving behaviors, demonstrating the synchronized temporal dynamics captured by the proposed preprocessing pipeline.

\subsection{Multimodal Feature Extraction}
We employed a domain-specific feature extraction pipeline targeting distinct physiological features within an architecture that processes three primary modalities comprising 59 EEG channels, 4 EMG channels, and a single GSR channel. The feature extraction strategy comprises three complementary categories:

\textbf{Time-Domain Features:} For each channel, we extracted metrics capturing local signal dynamics, including line length ($LL$), variance of the 1st derivative, and maximum absolute change. These features are robust to amplitude scaling and baseline shifts. Specifically, the line length, which quantifies the complexity and fractal dimension of the signal by measuring the cumulative distance between consecutive data points.

\begin{equation}
    LL = \sum_{i=1}^{N-1} |x[i+1] - x[i]|
    \label{eq:line_length}
\end{equation}

\noindent where $x[i]$ represents the value of the $i$-th sample in a given epoch, and $N$ denotes the total number of samples within that window.

\textbf{Frequency-Domain Features:} To quantify the distribution of signal power across different frequency components, the power spectral density (PSD) was estimated using Welch’s method. This technique improves the standard periodogram by averaging modified periodograms from overlapping segments, thereby reducing the variance of the estimator. 

\begin{equation}
    \hat{P}_{Welch}(f) = \frac{1}{K} \sum_{m=1}^{K} \left[ \frac{1}{N \cdot U} \left| \sum_{n=0}^{N-1} x_m[n] \cdot w[n] e^{-j 2 \pi f n} \right|^2 \right]
    \label{eq:welch}
\end{equation}

\noindent where $K$ represents the number of overlapping segments, $N$ is the length of each segment, and $x_m[n]$ denotes the $n$-th sample of the $m$-th segment. The term $w[n]$ refers to the window function (e.g., Hamming window) applied to each segment to minimize spectral leakage, and $U$ is a normalization factor given by $U = \frac{1}{N} \sum_{n=0}^{N-1} |w[n]|^2$. Finally, the computed PSD values were averaged within modality-specific frequency bands to serve as robust features for classification.

\begin{itemize}
    \item \textbf{EEG:} Delta (0.5-4 Hz), Theta (4-8 Hz), Alpha (8-13 Hz), Beta (13-30 Hz), and Gamma (30-50 Hz).
    \item \textbf{EMG:} Low (20-60 Hz), Mid (60-100 Hz), and High (100-240 Hz) frequency ranges.
    \item \textbf{GSR:} Phasic response and noise components.
\end{itemize}

\textbf{Derived Physiological Indices:} To capture inter-modality dynamics and global cognitive states, we computed global indicators such as the EEG Alpha/Theta Ratio (a proxy for cognitive load and drowsiness) and the EMG asymmetry index (quantifying muscular imbalance during steering maneuvers). 

This process generated an initial feature space of 503 dimensions. EEG contributed the majority of features due to its high channel count, while EMG and GSR provided critical motor and arousal indicators, respectively.

\subsection{SHAP-Based Feature Selection}
High-dimensional feature spaces pose risks of overfitting and increased computational overhead. To address this, we implemented a SHAP-based elite feature selection strategy. This approach is grounded in cooperative game theory, where each feature is treated as a player in a coalition, and its contribution to the model output is quantified using Shapley values. 

\begin{equation}
    \phi_i(f, x) = \sum_{S \subseteq \{x_1, \dots, x_p\} \setminus \{x_i\}} \frac{|S|!(p - |S| - 1)!}{p!} [f(S \cup \{x_i\}) - f(S)]
    \label{eq:shap_formula}
\end{equation}

where $\phi_i$ is the SHAP value for feature $i$, $p$ is the total number of features, the symbol $!$ denotes the factorial operation, $S$ represents a subset of features excluding $x_i$, and $f(S)$ denotes the expected model output given the features in set $S$. A lightweight XGBoost classifier was initially trained on the full 503-feature space. SHAP values were computed using TreeExplainer and aggregated across all $N$ samples and $C$ classes to generate a robust 1D importance vector, $I_j$.

\begin{equation}
    I_j = \frac{1}{N} \sum_{k=1}^{N} \sum_{c=1}^{C} |\phi_{j}^{(k,c)}|
    \label{eq:shap_agg}
\end{equation}

In this context, $I_j$ represents the global importance of the $j$-th feature, $N$ is the total number of samples, and $\phi_{j}^{(k,c)}$ is the SHAP value of feature $j$ for the $k$-th sample regarding class $c$. This aggregation ensures that feature importance reflects global predictive power rather than class-specific biases.

Based on this analysis, the top 250 unique features ($j \in \text{elite set}$ where $I_j$ is maximized) were selected for the final training phase. Preliminary experiments indicated that this subset retains approximately 81\% of the model's predictive power while significantly reducing dimensionality by $\approx 50\%$. Crucially, this selection process was performed exclusively on the training data to maintain the integrity of the test set evaluation.

\subsection{Hyperparameter Optimization}
To maximize the performance of the gradient boosting models, we employed Bayesian optimization via the tree-structured Parzen estimator (TPE) algorithm, implemented in the Optuna framework \cite{akiba2019optuna}. Separate optimization procedures were conducted for XGBoost and LightGBM. The search space included learning rates, tree depths, subsample ratios, and regularization parameters (L1/L2), as detailed in Table \ref{tab:hyperparams}. Additionally, the ensemble blending weight $\alpha$ (representing the contribution of XGBoost) was optimized within the range $[0.0, 1.0]$. Each configuration was evaluated using 5-fold stratified cross-validation on the training set, with the macro-F1 score as the optimization objective (maximized). macro-F1 was chosen to ensure balanced performance across all driving behavior classes, particularly minority classes like Throttle. The trial budget was set to 50 iterations per model, balancing computational cost with exploration depth.

The optimization process identified distinct optimal configurations for each model. Notably, the final ensemble utilized an optimized blending ratio of $\alpha = 0.35$ for XGBoost and $(1-\alpha) = 0.65$ for LightGBM. This weighting suggests that while both models contribute significantly, LightGBM’s leaf-wise growth strategy provides a slight edge in handling the SHAP-selected feature space. The use of TPE allowed for efficient exploration of the hyperparameter space, converging on configurations that favor deeper trees for XGBoost and conservative regularization for LightGBM, confirming their complementary modeling strengths.

\begin{table}[h]
\centering
\caption{Hyperparameter Search Space and Optimization Settings.}
\label{tab:hyperparams}
\footnotesize
\begin{tabular}{lcc}
\toprule
Parameter & XGBoost Range & LightGBM Range \\
\midrule
Number of Estimators & [500, 2000], step=100 & [500, 2000], step=100 \\
Learning Rate & [0.005, 0.05], log-scale & [0.005, 0.05], log-scale \\
Maximum Depth & [4, 10] & [4, 10] \\
Subsample Ratio & [0.6, 1.0] & [0.6, 1.0] \\
Feature Sampling (colsample) & [0.5, 1.0] & [0.5, 1.0] \\
Min Child Weight & [1, 10] & — \\
Gamma & [0.0, 1.0] & — \\
Max Delta Step & [0, 5] & — \\
Num Leaves & — & [20, 150] \\
Min Child Samples & — & [10, 100] \\
Regularization (L1) & — & [$10^{-4}$, 10.0], log-scale \\
Regularization (L2) & — & [$10^{-4}$, 10.0], log-scale \\
\midrule
Objective & multi:softprob & multiclass \\
CV Folds & 5 & 5 \\
Optimization Metric & macro-F1 & macro-F1 \\
Number of Trials & 50 & 50 \\
\bottomrule
\end{tabular}
\end{table}

\subsection{Ensemble Strategy}
The core predictive engine is a Soft-Voting Ensemble combining XGBoost and LightGBM. This hybrid approach leverages XGBoost's robustness to outliers and LightGBM's efficiency with high-dimensional data. The final classification decision is based on the weighted averaging of the predicted class probabilities.

\begin{equation}
    P_{final}(k|x) = \alpha \cdot P_{XGB}(k|x) + (1 - \alpha) \cdot P_{LGB}(k|x)
    \label{eq:soft_voting}
\end{equation}

where $P_{final}(k|x)$ represents the ensemble's predicted probability for class $k$ given input feature vector $x$. $P_{XGB}$ and $P_{LGB}$ denote the posterior probabilities generated by the XGBoost and LightGBM members, respectively. The ensemble weight $\alpha$ represents the contribution of XGBoost and was optimized alongside other hyperparameters. Our final configuration utilizes an optimized blending ratio of $\alpha = 0.35$ for XGBoost and $(1 - \alpha) = 0.65$ for LightGBM. This weighting balances the gradient-based learning strategies of both models, with LightGBM playing a dominant role in the final decision boundary, likely due to its superior handling of the sparse, SHAP-selected feature space. The final predicted class $\hat{y}$ is determined by selecting the category with the highest aggregated probability:
\begin{equation}
    \hat{y} = \arg\max_{k \in \{1, \dots, C\}} P_{final}(k|x)
\end{equation}

To address class imbalance, balanced sample weights $w_c$ were applied during training, inversely proportional to class frequency $n_c$:

\begin{equation}
    w_c = \frac{N}{C \cdot n_c}
\end{equation}

where $N$ is the total number of samples and $C$ is the number of classes. This ensures that minority categories receive adequate influence during gradient updates without the need for synthetic data generation.

\subsection{Evaluation Protocol}
Model performance was assessed on the held-out test set ($n=3,577$) using accuracy and macro-F1 score. Accuracy provides a global measure of correct classifications, defined as the ratio of correctly predicted samples to the total number of samples:

\begin{equation}
    \text{accuracy} = \frac{TP + TN}{TP + TN + FP + FN}
\end{equation}

where $TP$, $TN$, $FP$, and $FN$ denote true positives, true negatives, false positives, and false negatives, respectively. To ensure consistent performance across all classes, particularly for minority categories, the macro-F1 score was utilized. Unlike the micro-averaged version, the macro-F1 calculates the F1-score independently for each class and then takes the unweighted average, thereby giving equal importance to each driving behavior:

\begin{equation}
    \text{F1}_{class} = \frac{2 \cdot \text{Precision} \cdot \text{Recall}}{\text{Precision} + \text{Recall}}
\end{equation}

\begin{equation}
    \text{macro-F1} = \frac{1}{C} \sum_{i=1}^{C} \text{F1}_{i}
\end{equation}

where $C$ is the number of classes. This metric is particularly sensitive to the model's ability to generalize to underrepresented maneuvers in the MPDB dataset. To rigorously evaluate the proposed framework, the following experimental protocols were established:

\begin{enumerate}
    \item \textbf{Comprehensive Baseline Comparison:} To benchmark the proposed XGB+LGB ensemble, we implemented a suite of traditional machine learning classifiers, including random forest (RF with 500 estimators, max depth 25), support vector machine (SVM with RBF kernel), k-nearest neighbors (KNN with $k=15$), decision tree (DT), and multinomial logistic regression (MLR). All baselines were trained on the same SHAP-selected elite feature space to ensure a fair comparison.
    
    \item \textbf{Multi-level Ablation Study:} We conducted a systematic ablation analysis to quantify the necessity of multimodal fusion. This involved training independent models on:
    \begin{itemize}
        \item \textbf{Single-modality sets:} EEG-only, EMG-only, and GSR-only.
        \item \textbf{Pairwise-modality combinations:} EEG+EMG, EEG+GSR, and EMG+GSR to identify synergistic interactions between different physiological streams.
    \end{itemize}
    
    \item \textbf{Interpretability and Stability Analysis:} Beyond predictive metrics, the model's physiological plausibility was validated using SHAP. This included global feature importance rankings, modality-level contribution decomposition via donut charts, and local explanation analysis through SHAP interaction and dependence plots to ensure the model's decisions align with established neurophysiological principles.
\end{enumerate}

\section{Results}

\subsection{Global Classification Performance}
The proposed multimodal ensemble framework demonstrated robust performance, achieving an overall accuracy of 80.91\% and a macro-F1 score of 0.79 on the held-out test set. Table \ref{tab:model_comparison_new} presents a comprehensive comparison of the proposed ensemble against various baseline classifiers and ablation configurations. 

The proposed framework significantly outperformed the traditional RF baseline (accuracy: 72.99\%, macro-F1: 0.70), yielding an absolute improvement of approximately 7.92\% in accuracy. This performance gap highlights the efficacy of the SHAP-based feature selection strategy and the hybrid gradient boosting ensemble (XGBoost + LightGBM) in capturing complex, non-linear physiological patterns associated with driving behaviors. Furthermore, the ensemble surpassed other classical machine learning baselines, including KNN (64.91\%), MLR (56.05\%), SVM (53.56\%), and DT (48.40\%), confirming that gradient-boosted trees are particularly well-suited for this high-dimensional, structured physiological data.

\begin{figure}[h]
\begin{center}
\includegraphics[width = 16 cm, clip = true, trim = 0 0 0 0]{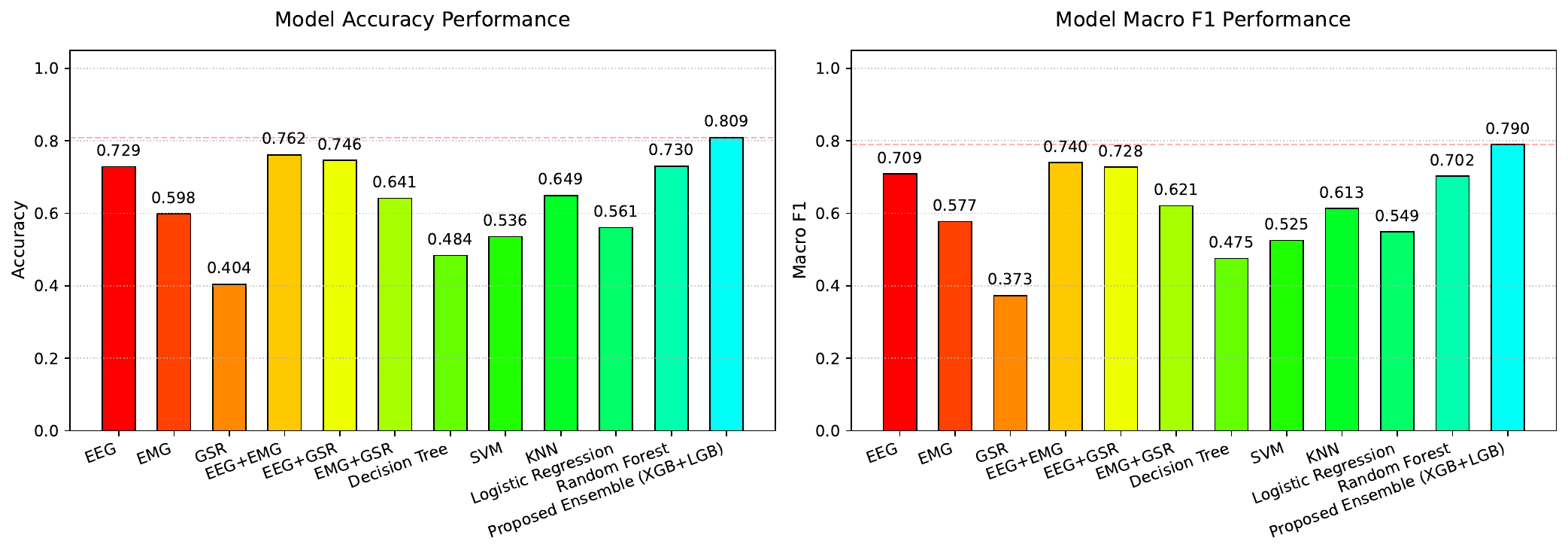}
\caption{Comparison of model performance across accuracy and macro-F1 metrics. The proposed ensemble (XGB+LGB) demonstrates clear superiority over baseline classifiers, single-modality, and pair-wise models, highlighting the necessity of multimodal data fusion and advanced ensemble learning for robust driving behavior classification.}
\label{fig:model_comparison}
\end{center}
\end{figure}

The superiority of the hybrid ensemble strategy is visually summarized in Figure \ref{fig:model_comparison}, which illustrates the consistent performance advantage of the proposed method across both accuracy and F1-score metrics compared to single-model baselines.

\begin{table}[h]
\centering
\caption{Comprehensive Model Performance Comparison and Ablation Analysis.}
\label{tab:model_comparison_new}
\footnotesize
\begin{tabular}{lccc}
\toprule
Model & Accuracy & Macro-F1 & Performance Gap (Acc.) \\
\midrule
\textbf{Proposed Ensemble (XGB+LGB)} & \textbf{80.91\%} & \textbf{0.79} & — \\
Random Forest             & 72.99\%          & 0.70          & -7.92\% \\
KNN                       & 64.91\%          & 0.61          & -16.00\% \\
Logistic Regression       & 56.05\%          & 0.54          & -24.86\% \\
SVM                       & 53.56\%          & 0.52          & -27.35\% \\
Decision Tree             & 48.40\%          & 0.47          & -32.51\% \\

\midrule
\multicolumn{4}{l}{\textit{Comprehensive Modality Ablation}} \\
EEG + EMG                            & 76.15\%          & 0.74          & -4.67\% \\
EEG + GSR                            & 74.59\%          & 0.73          & -6.32\% \\
EMG + GSR                            & 64.14\%          & 0.62          & -16.77\% \\
EEG Only                             & 72.85\%          & 0.71          & -8.06\% \\
EMG Only                             & 59.83\%          & 0.57          & -21.08\% \\
GSR Only                             & 40.59\%          & 0.37          & -40.32\% \\
\bottomrule
\end{tabular}
\end{table}

\begin{figure}[h]
\begin{center}
\includegraphics[width = 10 cm, clip = true, trim = 0 0 0 0]{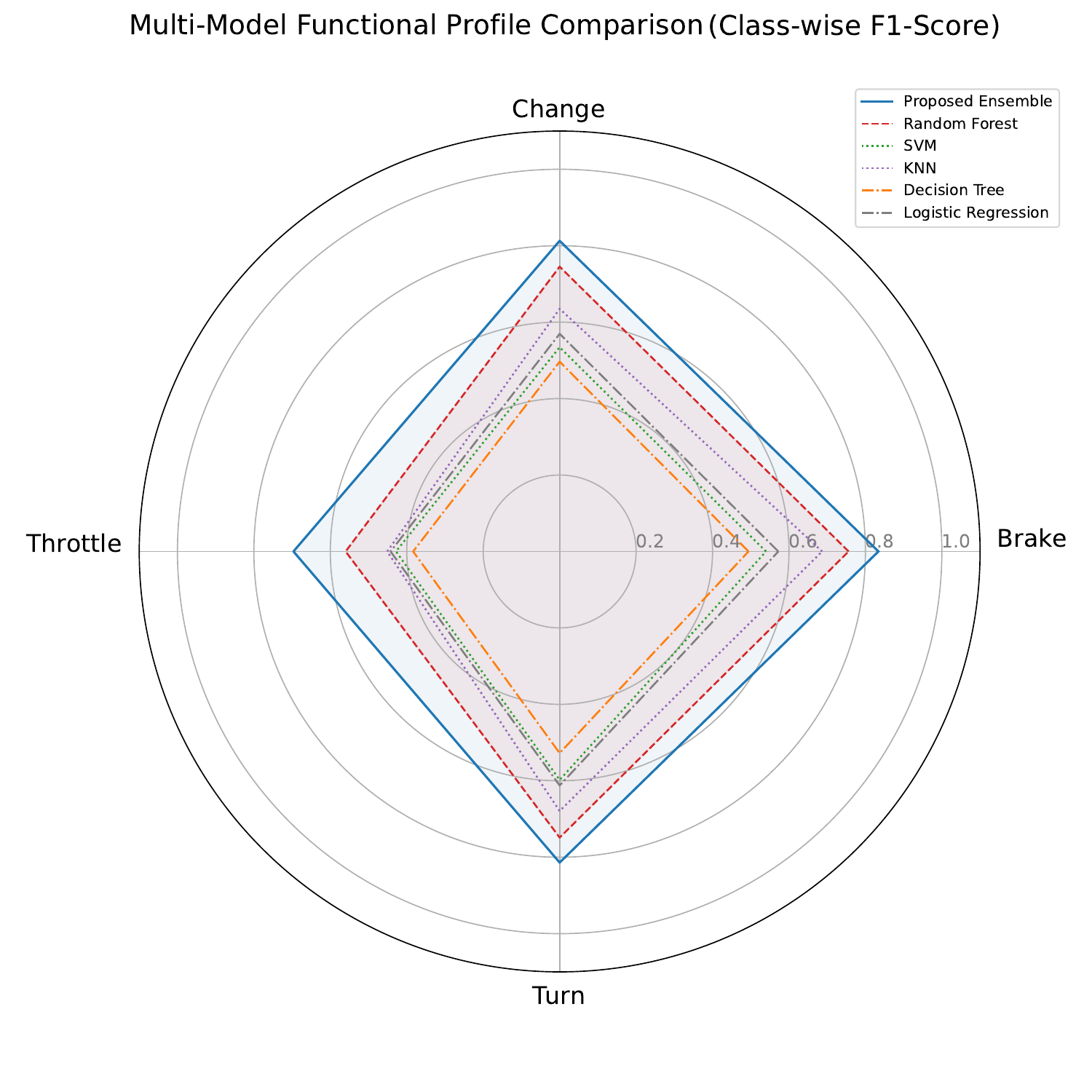}
\caption{Functional performance profile comparison across the four targeted driving behaviors. The radar chart illustrates the F1-score of the proposed XGB+LGB ensemble against various baseline models, including RF, SVM, KNN, DT, and MLR. The proposed ensemble demonstrates superior and more balanced discriminative power across all classes.}
\label{fig:comprehensive_radar_comparison}
\end{center}
\end{figure}

To further evaluate the robustness of the proposed framework, Figure \ref{fig:comprehensive_radar_comparison} illustrates a multi-model functional profile analysis conducted via a radar chart. This visualization compares the F1-scores of the XGB+LGB ensemble against traditional machine learning baselines across the four behavior classes. While simpler models like MLR and DT show significant performance degradation in minority or subtle classes such as Throttle, the proposed ensemble maintains high and consistent F1-scores across all maneuvers. Notably, the ensemble's polygon almost entirely encompasses the profiles of other models, confirming its superior generalization and its ability to effectively leverage multimodal physiological cues for diverse driving behaviors.

\subsubsection{Per-Class Performance Analysis}
Detailed per-class metrics, presented in Table \ref{tab:per_class_new}, reveal heterogeneous but generally strong classification quality across the four behavior categories. The model exhibited distinct proficiency in identifying high-arousal and high-motor-demand maneuvers:

\begin{itemize}
    \item \textbf{Brake} and \textbf{Turn} behaviors achieved the highest reliability, with F1-scores of 0.83 and 0.81, respectively. The multimodal fusion effectively captures the distinct physiological signatures of these maneuvers: acute sympathetic arousal for braking, and asymmetric muscular activation (detectable via EMG) for turning.
    \item \textbf{Change} (Lane Change) showed strong performance (F1: 0.79), benefiting from the combination of cognitive attention shifts (EEG) and lateral motor preparation (EMG).
    \item \textbf{Throttle} (Acceleration) remained the most challenging class (F1: 0.69). This is likely due to the subtler physiological manifestation of smooth acceleration compared to the abrupt changes seen in braking or turning. Acceleration often involves gradual increases in cognitive load and muscle tension, leading to higher confusion rates.
\end{itemize}

\begin{table}[h]
\centering
\caption{Detailed Per-Class Classification Performance of the Proposed Ensemble.}
\label{tab:per_class_new}
\footnotesize
\begin{tabular}{lcccc}
\toprule
Class & Precision & Recall & F1-Score & Support \\
\midrule
Brake    & 0.81 & 0.86 & 0.83 & 1,283 \\
Change   & 0.81 & 0.77 & 0.79 & 627 \\
Throttle & 0.77 & 0.63 & 0.69 & 489 \\
Turn     & 0.80 & 0.82 & 0.81 & 1,178 \\
\midrule
Accuracy        & \multicolumn{3}{c}{0.80} & 3,577 \\
Macro Average    & 0.80 & 0.77 & 0.78 & 3,577 \\
Weighted Average & 0.80 & 0.80 & 0.80 & 3,577 \\
\bottomrule
\end{tabular}
\end{table}

\subsection{Modality Contribution and Ablation Study}
The ablation analysis confirms the critical necessity of the multimodal approach. As detailed in Table \ref{tab:model_comparison_new}, the removal of any modality results in performance degradation, though the extent varies:

\begin{itemize}
    \item \textbf{EEG Only:} Achieved 72.85\% accuracy. EEG serves as the strongest single modality, providing high sensitivity to cognitive load, attention shifts, and mental workload associated with decision-making processes.
    \item \textbf{EMG Only:} Achieved 59.83\% accuracy. While effective for capturing explicit motor responses (steering angle, pedal pressure), EMG lacks the contextual cognitive state information required to distinguish between mentally distinct but physically similar maneuvers.
    \item \textbf{GSR Only:} Achieved the lowest standalone performance (40.59\% accuracy). This is attributed to the slow temporal response of skin conductance, which struggles to resolve rapid, transient driving events within short 2-second windows.
    \item \textbf{Pairwise Fusion:} Combining EEG with EMG yielded the best pairwise result (76.15\% accuracy), closely approaching the full multimodal performance. This suggests that the synergy between cognitive (EEG) and motor (EMG) signals is the primary driver of classification accuracy. However, the full ensemble (80.91\%) still provides a notable ~4.7\% gain over EEG+EMG, indicating that GSR and other subtle signals contribute unique, non-redundant information, particularly for distinguishing high-arousal states like braking.
\end{itemize}

Critically, the proposed multimodal framework results in an ~8\% improvement in accuracy over the best single-modality model (EEG). This synergy validates that EEG, EMG, and GSR provide complementary information essential for robust driver monitoring.

\subsection{SHAP Analysis and Interpretability}
SHAP analysis provided granular insights into the biological drivers of the model's decisions, ensuring that the high performance is grounded in physiological plausibility.

\begin{figure}[h]
\begin{center}
\includegraphics[width = 7 cm, clip = true, trim = 0 0 0 0]{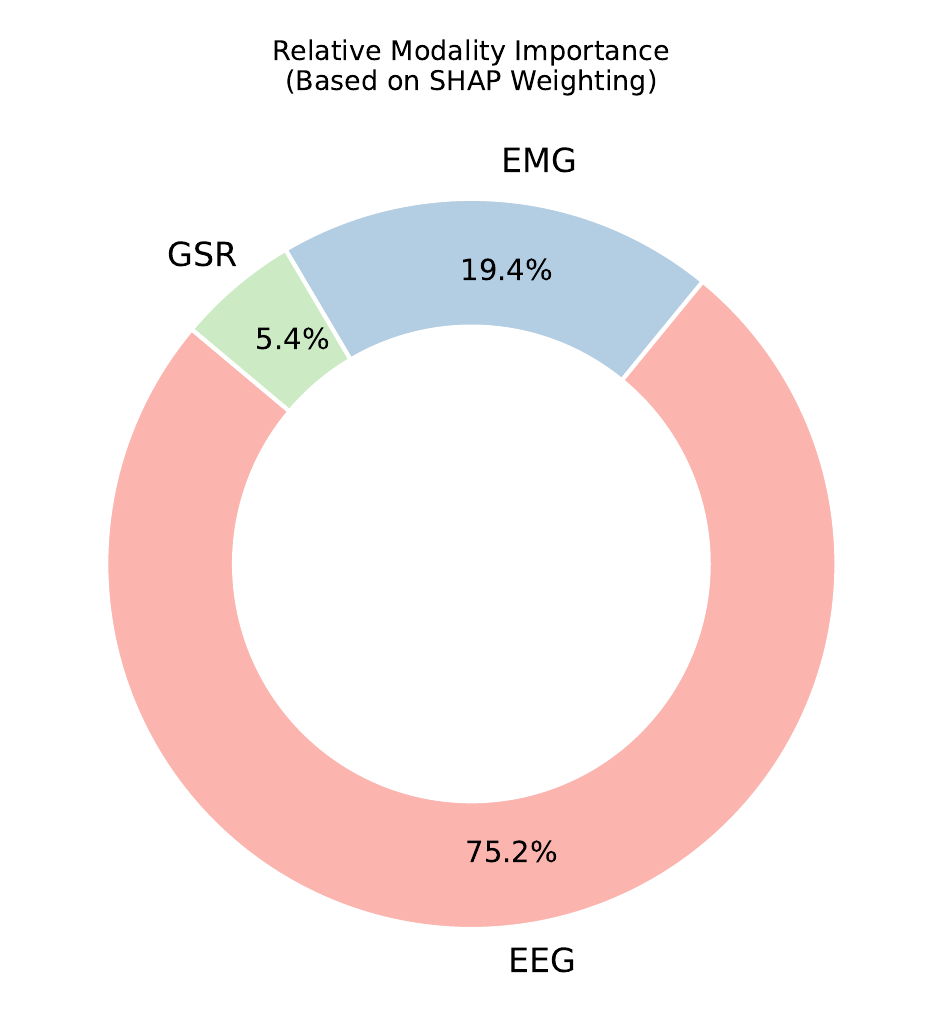}
\caption{Relative Modality Importance based on cumulative SHAP weighting. Although individual GSR and EMG features rank highly in discriminative power, the cumulative weight of multi-channel EEG (75.2\%) underscores the central role of cognitive state monitoring in the overall decision process.}
\label{fig:modality_donut_chart}
\end{center}
\end{figure}

\begin{figure}[h]
\begin{center}
\includegraphics[width = 14 cm, clip = true, trim = 0 0 0 0]{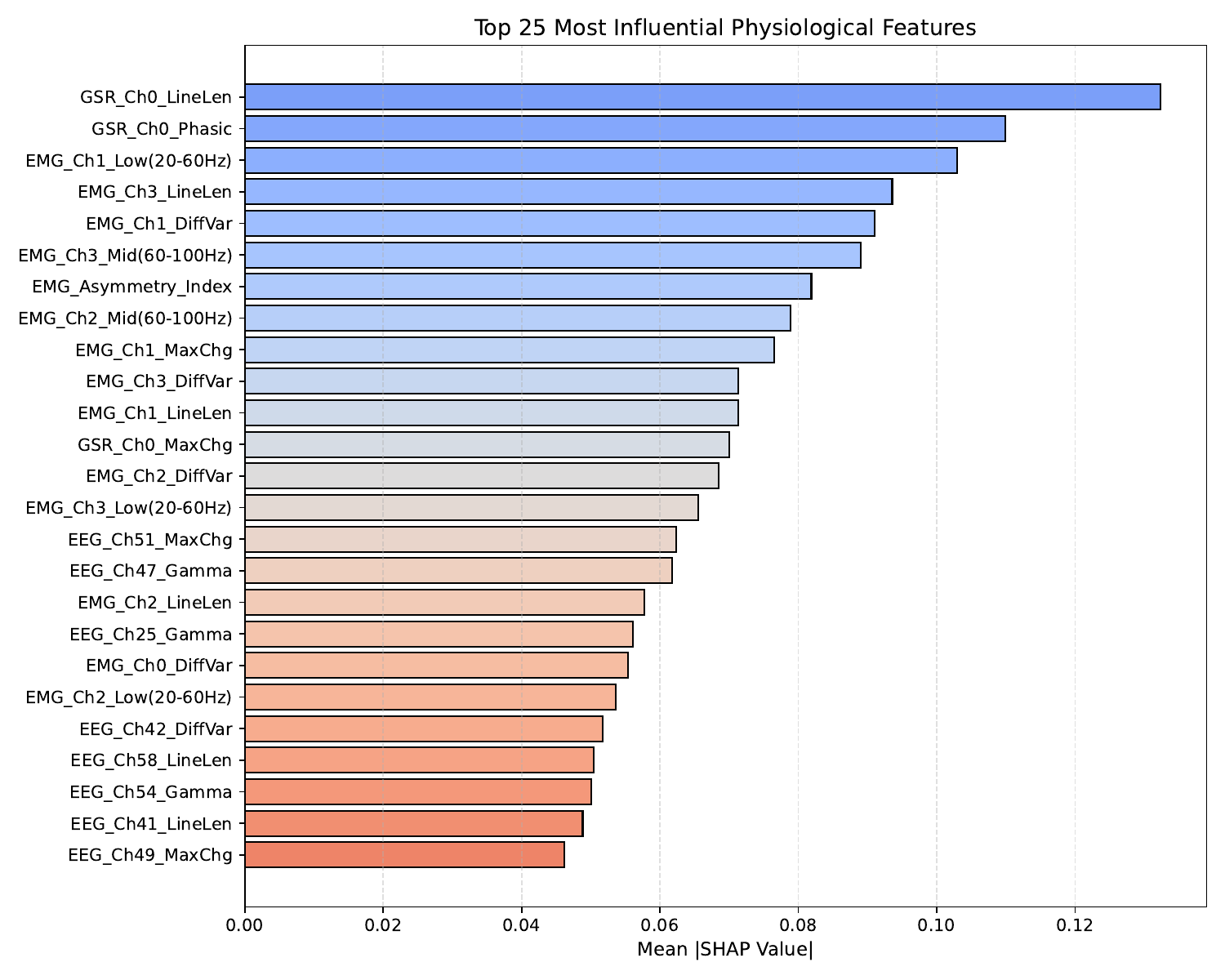}
\caption{Top 25 Most Influential Physiological Features based on Mean SHAP Value. GSR-based temporal features and EMG spectral components emerge as critical individual predictors, complementing the broader cognitive context provided by EEG.}
\label{fig:shap_top25}
\end{center}
\end{figure}

\begin{itemize}
    \item \textbf{Modality Weight:} As illustrated in Figure \ref{fig:modality_donut_chart}, EEG contributes the largest share (75.2\%) to the total model decision weight, followed by EMG (19.4\%) and GSR (5.4\%). This distribution reflects both the higher dimensionality of EEG features (59 channels) and the central role of cognitive processing in anticipating and executing driving maneuvers.
    
    \item \textbf{Top Discriminative Features:} Despite the aggregate dominance of EEG, the individual most influential features often originated from peripheral signals. As shown in Figure \ref{fig:shap_top25}, GSR-based temporal features (line length and phasic components) and EMG spectral components (low-frequency power and asymmetry index) ranked among the top predictors. 
    \begin{itemize}
        \item The prominence of GSR line length and phasic activity aligns with their sensitivity to acute sympathetic arousal, making them critical for detecting \textit{Brake} events.
        \item The EMG asymmetry index emerged as a key discriminator for \textit{Turn} maneuvers, reflecting the biomechanical reality of differential muscle activation during steering.
    \end{itemize}
\end{itemize}

These findings suggest a hierarchical contribution: while EEG provides the broad structural context of the driver's cognitive state, GSR and EMG provide sharp, high-contrast signals for specific high-arousal or high-motor events. The interpretability analysis thus validates the model, linking high-importance features directly to established physiological markers of driving behavior.

As illustrated in Table~\ref{tab:comprehensive_comparison}, our proposed XGB+LGB framework significantly outperforms existing benchmarks on the MPDB dataset, achieving a state-of-the-art accuracy of 80.91\%. While the foundational study by Tao et al. established a peak accuracy of 66.2\% with the multimodal LSTM architecture, and Zhang et al. further enhanced performance to 74.40\% through the use of hypergraph convolutional networks, our approach leverages SHAP elite feature selection and Infomax ICA to better capture relevant physiological patterns. These results demonstrate that the combination of gradient-boosted decision trees and refined feature engineering provides a more robust solution for classifying complex driving behaviors compared to traditional deep learning and sequential models. Since machine learning applications to this dataset remain relatively limited, the available comparative baselines are still narrow. This highlights an opportunity for future studies to explore more sophisticated learning algorithms and broader experimental settings.

\begin{table}[H]
\centering
\caption{Comprehensive performance comparison on the MPDB dataset.}
\label{tab:comprehensive_comparison}
\footnotesize
\begin{tabular}{lllcc} 
\toprule
Study & Model & Modalities & Accuracy & F1-score \\ \midrule
\multirow{7}{*}{Tao et al. \cite{tao2024multimodal}} & LDA & EEG Only & 33.50\% & — \\
 & LDA & EEG, EMG, GSR, ECG & 36.90\% & — \\
 & EEGNet & EEG Only & 49.80\% & — \\
 & EEGNet & EEG, EMG, GSR, ECG & 57.70\% & — \\
 & CNN & EEG, EMG, GSR, ECG & 62.50\% & — \\
 & MMPNet & EEG, EMG, GSR, ECG & 65.30\% & — \\
 & LSTM & EEG, EMG, GSR, ECG & 66.20\% & — \\ \midrule
\multirow{7}{*}{Zhang et al. \cite{zhang2024mmphgcn}} & KNN & EEG, EMG, GSR & 48.42\% & 42.57\% \\
 & EEGNet & EEG, EMG, GSR & 55.71\% & 48.54\% \\
 & C-TSF & EEG, EMG, GSR & 56.60\% & 53.03\% \\
 & 1DCNN & EEG, EMG, GSR & 63.14\% & 57.68\% \\
 & M-mixer & EEG, EMG, GSR & 66.69\% & 61.89\% \\
 & H-LSTM & EEG, EMG, GSR & 68.51\% & 63.26\% \\
 & MMPHGCN & EEG, EMG, GSR & 74.40\% & 70.64\% \\ \midrule
\textbf{Ours} & \textbf{XGB+LGB} & \textbf{EEG, EMG, GSR} & \textbf{80.91\%} & \textbf{79.00\%} \\ \bottomrule
\end{tabular}
\end{table}

\section{Discussion}

The results of this study demonstrate that a rigorously engineered pipeline can effectively decode complex driving behaviors from multimodal physiological signals by combining advanced signal preprocessing, SHAP-based elite feature selection, and a hybrid gradient boosting ensemble. The proposed framework achieved a test accuracy of 80.91\% and a macro-F1 score of 0.79 on the large-scale MPDB dataset. This performance significantly outperforms both single-modality baselines and traditional machine learning models, validating the efficacy of the multimodal fusion strategy. This section discusses the implications of these findings, the physiological interpretability of the model, and the limitations of the current approach.

\subsection{Effectiveness of Multimodal Fusion and Ensemble Learning}
The substantial performance gap between the proposed multimodal ensemble and single-modality models (Table \ref{tab:model_comparison_new}) underscores the complementary nature of physiological signals in driving behavior analysis. While EEG alone achieved the highest single-modality performance (72.85\% accuracy), it failed to capture the muscular dynamics critical for distinguishing maneuvers like turning. Conversely, EMG provided strong signals for physical actions (59.83\% accuracy) but lacked the cognitive context provided by EEG. GSR, while weak in isolation (40.59\% accuracy), contributed critical arousal markers when fused. The integration of these modalities allowed the model to construct a holistic representation of the driver's state, integrating cognitive load, autonomic arousal, and motor execution. The 8\% improvement in accuracy over the best single modality (EEG) confirms that no single physiological stream is sufficient for robust classification of dynamic driving events.

Furthermore, the superiority of the XGBoost-LightGBM ensemble over the RF baseline (80.91\% vs. 72.99\% accuracy) highlights the importance of gradient-based optimization in handling the non-linear and high-dimensional nature of physiological features. The weighted soft-voting strategy, which favored LightGBM ($\alpha=0.65$ for LGB, $\alpha=0.35$ for XGB), suggests that LightGBM’s leaf-wise growth strategy and histogram-based splitting were particularly effective in capturing the subtle interactions within the SHAP-selected feature space. The consistent performance across the large, stratified test set ($n=3,577$) further confirms that this performance is robust and generalizable, not an artifact of specific data splits or small sample sizes.

\subsection{Physiological Interpretability via SHAP Analysis}
A key contribution of this work is the integration of SHAP values to enhance model transparency. Unlike deep learning black boxes, our tree-based approach allows for direct mapping of model decisions to physiological phenomena. 

\textbf{Modality Contribution:} SHAP decomposition revealed that EEG contributes the largest share of predictive value (75.2\%), followed by EMG (19.4\%) and GSR (5.4\%). This distribution reflects both the higher dimensionality of EEG features (59 channels) and the central role of cognitive processes in driving behavior. However, the non-zero contributions from all modalities confirm that each sensor provides unique information not redundant with others.

\textbf{Feature-Level Insights:} Despite EEG's dominance in total weight, the most influential individual features included GSR Line Length and GSR Phasic components, followed by EMG low-frequency (20-60 Hz) power and the EMG asymmetry index. 
\begin{itemize}
    \item \textbf{Cognitive Load:} The prominence of EEG-derived features, particularly in the Alpha and Theta bands, aligns with established literature linking these oscillations to visual attention and mental workload \cite{klimesch2012alpha,cavanagh2014frontal}. The model correctly identifies changes in cognitive state associated with different driving maneuvers.
    \item \textbf{Motor Execution:} The significant contribution of the EMG asymmetry index for classifying turns provides biomechanical validation. Turning requires differential activation of left and right upper-body muscles to steer, a pattern distinctly captured by asymmetry features.
    \item \textbf{Autonomic Arousal:} The high importance of GSR phasic components suggests that sympathetic nervous system activation is a critical discriminator for high-arousal maneuvers like braking, even if GSR's overall contribution is lower due to its slower temporal dynamics.
\end{itemize}
These findings confirm that the model is not merely exploiting statistical artifacts but is learning physiologically plausible representations of driver behavior.

\subsection{Challenges in Class Discrimination}
Despite the overall strong performance, the model exhibited heterogeneous performance across classes. Brake and Turn behaviors showed the highest reliability (F1 scores of 0.83 and 0.81, respectively). These maneuvers involve distinct physiological signatures: acute sympathetic arousal and cognitive focus for braking, and asymmetric muscular activation for turning.

In contrast, the Throttle class proved the most challenging (F1 = 0.69). This difficulty likely stems from the subtle physiological manifestation of acceleration compared to active maneuvers like braking or turning. Acceleration may not always trigger the same level of acute sympathetic arousal or distinct motor patterns, leading to greater overlap with baseline or smooth driving states. Additionally, the variability in how drivers execute acceleration (gradual vs. aggressive) may introduce noise that is harder to resolve with fixed 2-second windows. Future work could explore finer-grained labeling of acceleration intensity or the inclusion of vehicle kinematic data to help disambiguate these overlapping physiological patterns.

\subsection{Discussion of Future Trajectories}
While the proposed framework establishes a robust and interpretable baseline for physiological driver monitoring, several avenues for further enhancement remain. The current study utilizes a high-fidelity simulated driving dataset, which facilitates precise event synchronization and artifact control. Future work will aim to validate these findings under diverse real-world traffic conditions to further explore the influence of dynamic environmental stressors on physiological responses. 

Furthermore, while the fixed 2-second windowing strategy proved computationally efficient and well-aligned with behavioral onsets, investigating adaptive temporal windows or integrating recurrent or transformer-based architectures could offer deeper insights into long-term signal dependencies. Lastly, to transition from the current scalable streaming pipeline to real-time in-vehicle deployment, future iterations will focus on model compression techniques, such as quantization and pruning, to ensure high-speed inference on resource-constrained automotive edge devices. These directions will build upon the current framework to deliver more resilient and context-aware driver assistance systems.

\section{Conclusion}
This paper presents an interpretable and scalable framework for classifying driving behavior using multimodal physiological signals. By leveraging rigorous preprocessing (including ICA-based artifact removal and multimodal synchronization), streaming feature extraction, and SHAP-based elite feature selection, we addressed key challenges in biosignal processing, including inter-subject variability, noise, and high dimensionality. The proposed ensemble of XGBoost and LightGBM models achieved a test accuracy of 80.91\% and a macro-F1 score of 0.79 on the large-scale MPDB dataset, significantly outperforming single-modality baselines and traditional machine learning approaches.

Our results validate the hypothesis that multimodal physiological fusion, when combined with explainable machine learning, can provide robust and transparent insights into driver behavior. The SHAP analysis confirmed that the model relies on physiologically meaningful features, such as EEG alpha/theta ratios for cognitive load, EMG asymmetry for motor actions, and GSR phasic components for arousal, thereby enhancing trust in the system’s decisions. While challenges remain in distinguishing subtle maneuvers like throttle application, the framework offers a promising pathway toward real-time, uncertainty-aware driver monitoring systems. Future work will focus on extending this approach to driving data in uncontrolled environments, incorporating vehicle kinematics for context-aware classification, and optimizing the pipeline for embedded automotive hardware.

\section*{Data Availability Statement}

The MPDB \cite{tao2023mpdbdataset} dataset utilized in this study is a large-scale, publicly available resource comprising synchronized EEG, EMG, and GSR signals. The raw data and relevant documentation can be accessed via the official repository at \url{https://figshare.com/articles/dataset/Driving_behaviour_multimodal_human_factors_raw_dataset/22193119}

\section*{Declaration of Competing Interest}

The authors declare that they have no known competing financial interests or personal relationships that could have appeared to influence the work reported in this paper.

\section*{Authorship Contribution Statement}

\textbf{Sahar Askari:} Investigation, Methodology, Software, Writing – Original Draft, Writing – Review \& Editing. \textbf{Mohammad Mahdi Mirza Ali Mohammadi:} Investigation, Methodology, Software, Writing – Original Draft, Writing – Review \& Editing. \textbf{Fatemeh Ensafdoust:} Investigation, Writing – Original Draft, Writing – Review \& Editing. \textbf{Amin Golnari:} Validation, Formal Analysis, Investigation, Software, Visualization, Writing – Original Draft, Writing – Review \& Editing.  \textbf{Saeid Sanei:} Formal Analysis, Writing – Review \& Editing.

\bibliographystyle{ieeetr}  
\bibliography{references}

\end{document}